%% file: main.tex
\documentclass[sigconf]{acmart}
\AtBeginDocument{%
  \providecommand\BibTeX{{%
    \normalfont B\kern-0.5em{\scshape i\kern-0.25em b}\kern-0.8em\TeX}}}

\setcopyright{acmcopyright}
\copyrightyear{2020}
\acmYear{2020}
\setcopyright{iw3c2w3}
\acmConference[WWW '20]{Proceedings of The Web Conference 2020}{April 20--24, 2020}{Taipei, Taiwan}
\acmBooktitle{Proceedings of The Web Conference 2020 (WWW '20), April 20--24, 2020, Taipei, Taiwan}
\acmPrice{}
\acmDOI{10.1145/3366423.3380155}
\acmISBN{978-1-4503-7023-3/20/04}

\usepackage{booktabs} 
\usepackage{subfigure}
\usepackage{bm}

\usepackage{multirow}
\usepackage{algorithm}
\usepackage{algorithmic}
\usepackage{amssymb}
\usepackage{array}
\usepackage{graphicx}
\usepackage{listings}
\usepackage[np, autolanguage]{numprint}
\usepackage{enumitem}

\usepackage{soul}
\usepackage[skip=0pt]{caption}
\usepackage{subfigure}
\usepackage{amsmath}

\usepackage{xspace}
\newcommand{\paratitle}[1]{\vspace{1.5ex}\noindent\textbf{#1}}
\newcommand{\ie}{\emph{i.e.,}\xspace}
\newcommand{\aka}{\emph{a.k.a.,}\xspace}
\newcommand{\eg}{\emph{e.g.,}\xspace}

\newcommand{\ignore}[1]{}

\newcommand{\tabincell}[2]{\begin{tabular}{@{}#1@{}}#2\end{tabular}}
\begin{document}

\title{Mining Implicit Entity Preference from User-Item Interaction Data for Knowledge Graph Completion via Adversarial Learning}






\author{Gaole He$^{1,3}$, Junyi Li$^{2}$, Wayne Xin Zhao$^{2,3*}$, Peiju Liu$^{4}$ and Ji-Rong Wen$^{2,3}$}\thanks{$^*$Corresponding author.}
\affiliation{%
 \institution{$^1$School of Information, Renmin University of China}
 \institution{$^2$Gaoling School of Artificial Intelligence, Renmin University of China}
 \institution{$^3$Beijing Key Laboratory of Big Data Management and Analysis Methods}
 \institution{$^4$School of Electronics Engineering and Computer Science, Peking University}
}
\affiliation{%
  \institution{\{hegaole, lijunyi, jrwen\}@ruc.edu.cn, batmanfly@gmail.com, liupage2016@pku.edu.cn}
}	

\renewcommand{\shortauthors}{Gaole He, Junyi Li, Wayne Xin Zhao, Peiju Liu and Ji-Rong Wen}



\begin{abstract}
The task of Knowledge Graph Completion~(KGC) aims to automatically infer the missing fact information in Knowledge Graph (KG). 
In this paper, we take a new perspective that aims to  leverage rich user-item interaction data (\emph{user interaction data} for short) for improving the KGC task.
Our work is inspired by the observation that many KG entities correspond to online items in application systems. 
However, the two kinds of data sources have very different intrinsic characteristics, and it is likely to hurt the original  performance using simple fusion strategy.

To address this challenge, we propose a novel adversarial learning approach by leveraging user interaction data for the KGC task.  Our generator is  isolated from user interaction data, and serves to improve the performance of the discriminator.  The discriminator takes the learned useful information from user interaction data as input, and gradually enhances the evaluation capacity in order to identify the fake samples generated by the generator.
To discover implicit entity preference of users, we  design an elaborate collaborative learning algorithms based on graph neural networks, which will be jointly optimized with the discriminator.
Such an approach is effective to alleviate the issues about data heterogeneity and semantic complexity for the KGC task.
Extensive experiments on three real-world datasets have demonstrated the effectiveness of our approach on the KGC task. The source code is publicly available at \textcolor{blue}{https://github.com/RUCAIBox/UPGAN}.
\end{abstract}

\ccsdesc[300]{Computing methodologies~Knowledge representation and reasoning}

\keywords{Knowledge Graph, Adversarial Learning, User Preference}



\maketitle

\input{sec-intro}

\input{sec-rel}

\input{sec-pre}

\input{sec-model}

\input{sec-exp}
\input{sec-con}

\section*{Acknowledgement}
This work was partially supported by the National Natural Science Foundation of China under Grant No. 61872369 and 61832017, the Fundamental Research Funds for the Central Universities, the Research Funds of Renmin University of China under Grant No. 18XNLG22 and 19XNQ047, and Beijing Outstanding Young Scientist Program under Grant No. BJJWZYJH012019100020098, and Beijing Academy of Artificial Intelligence~(BAAI).  Xin Zhao is the corresponding author.
\bibliographystyle{ACM-Reference-Format}
\bibliography{www}

\end{document}

%% file: sec-intro.tex
\section{Introduction}


Recent years have witnessed the great thrive and wide application of large-scale knowledge graph~(KG).
 Although many existing KGs~\cite{yago,dbpedia,freebase,satori} are able to provide billions of structural facts about entities, they are known to be far from complete~\cite{GalarragaRAS17}. Hence,  various methods have been proposed to focus on the task of knowledge graph completion (KGC)~\cite{Bordes-NIPS-2013,Yang-CORR-2014,Dettmers-AAAI-2018}. Typically, KG represents a fact as a triple consisting of head entity, relation, tail entity.
 Based on this data form, the KGC task is usually described as predicting a missing entity in an incomplete triple. 

\begin{figure*}[!tb]
	\small
	\centering 
	\includegraphics[width=0.9\textwidth]{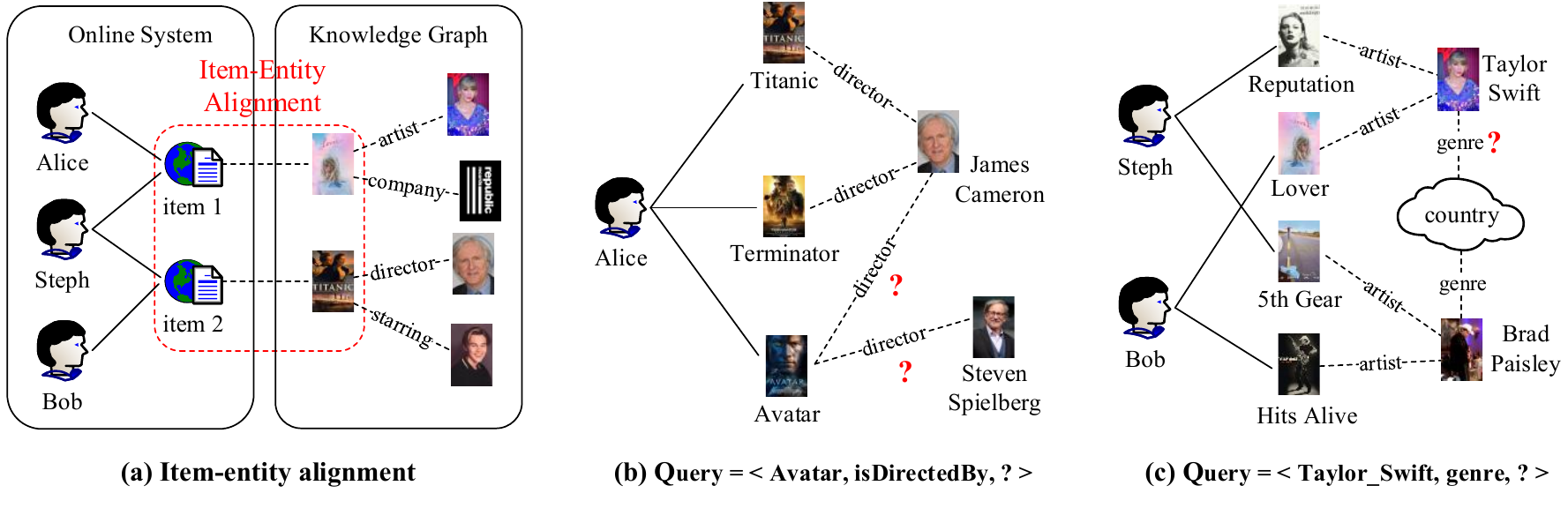} 
	\caption{Illustrative examples for our work: (a) item-entity alignment across online systems and KG entities in  movie and music domain; (b) inferring the director for the movie ``\emph{Avatar}''; and (c) inferring the artist genre for ``\emph{Taylor Swift}''.} 
	\label{fig-example} 
\end{figure*}

Most of previous KGC methods aim to devise new learning algorithms to  reason about underlying KG semantics using known fact information. 
In this work, we take a different perspective for tackling the KGC task.  
Since KG has been widely used in various applications, can we  leverage the accumulated application data for improving the KGC task? Specially, we are inspired by the observation that many KG entities correspond to online items in application systems. As shown in \cite{Zhao-DI-2019,Zhao-PAKDD-2019}, the items (\ie movies) from \textsc{MovieLens} have largely overlapped with the KG entities in Freebase. 
For KG entities aligned to online items,  we can obtain   fact triples from the KG as well as rich user-item interaction data (called \emph{user interaction data} for short) from the application platforms (See Fig.~\ref{fig-example}(a)). 
Based on this observation, the focus of this work is to study how user interaction data can be utilized to 
improve the KGC task. 

\ignore{Despite the fact that knowledge graph (KG) provide billions of machine-readable facts about entities, they are far from complete~\cite{}, and a dedicated line of research has focused on the task of knowledge graph completion (KGC)~\cite{}. Most previous studies that exploit KG-enabled features were based on the existing knowledge of KG, and focused on leveraging knowledge in one direction, i.e., from KG to other tasks. With the rapid development of Web techniques, recommendation systems (RS) play a more and more important role in matching user needs with rich resources (call items) from various online platforms. We are inspired by the fact that many KG entities correspond to online items in RS. As shown in \cite{Zhao-DI-2019}, the items (\ie movies) from \textsc{MovieLens} have largely overlapped with the KG entities in Freebase. Therefore, it is interesting that transfer the user-item interactions in RS to the KGC task with respect to the domain of items.}

User interaction data has explicitly reflected users' preference at the item level, while it is likely to contain implicit evidence about entity semantics, which is potentially useful to our task. Here, we present two illustrative examples. In Fig.~\ref{fig-example}(b), the user ``\emph{Alice}'' has watched three movies of ``\emph{Terminator}'', ``\emph{Titanic}'' and ``\emph{Avatar}'', and she is  a fan for the director of ``\emph{James Cameron}''. Given a query about the director of ``\emph{Avatar}'' and two candidate directors ``\emph{James Cameron}'' and ``\emph{Steven Allan Spielberg}'', knowing the user's interaction history is useful to identify the correct director in this case. As another example in music domain (See Fig.~\ref{fig-example}(c)), the users of ``\emph{Steph}'' and ``\emph{Bob}'' like the songs from both singers ``\emph{Taylor Swift}'' and ``\emph{Brad Paisley}'' due to the similar style. Such 
co-occurrence patterns in user interaction data are helpful to infer whether the two singers share the same artist genre in KG.
From two examples, it can be seen that user interaction data may contain useful preference information of users over KG entities.

Indeed, several recent efforts have attempted to leverage both KG data and user interaction data 
for jointly improving the KGC task and related recommendation tasks, including path-based methods~\cite{Sun-2018-RecSys}, regularization-based methods~\cite{Piao-2018-ESWC,Zhang-2018-Algorithms} and graph neural network methods~\cite{Wang-2019-KDD}. 
These studies mainly focus on developing data fusion models for integrating the two  kinds of data sources, \eg learning representations in the same space or share the same information representation across different sources. 
However, the two kinds of data sources have very different intrinsic characteristics, and it is likely to hurt the original representation performance using simple fusion strategy. 
In addition,  user interaction data is usually very noisy since user behaviors will be affected by external events (\eg on sale) or other influencing factors (\eg popularity). 
It may be problematic to  directly incorporate the learned information (\eg user preference) for inferring KG facts. 
To solve our task, we have to consider the effect of data heterogeneity and semantic complexity on model design.  
The major challenge can be summarized as: (1) how to learn useful information from user interaction data for improving KGC task and (2) how to integrate or utilize the learned information in  KGC methods.

As shown in Fig.~\ref{fig-example}, we can see that implicit entity preference of users is helpful to infer the plausibility of KG facts. 
Based on this motivation, our idea is to develop a specific evaluation component that incorporates and learns user preference information about entities for evaluating a candidate entity given a query~(\ie the head entity and relation).
Meanwhile, we keep a prediction component to produce the candidate entity without using user preference information.  
Since the prediction component tries to pass the check of the evaluation component by producing high-quality answers, it will tune and improve itself according to the feedback of the evaluation component. 
The two components will be improved via a mutual reinforcement process. 
By mapping the two components to discriminator and generator respectively, our idea naturally fits into the  successful framework of generative adversarial nets~(GAN)~\cite{GAN-2014-NIPS}.
In our setting, the discriminator is expected to effectively integrate the two kinds of heterogeneous data signals for the KGC task. While, the generator is  employed to improve the discriminator by modeling a pure KG semantic space. 


To this end, we propose a novel adversarial learning approach for leveraging user interaction data for the KGC task, named as \emph{UPGAN} (User Preference enhanced GAN). The proposed approach contains three major technical extensions. 
First, to learn useful evidence from user interaction data,  we integrate the two kinds of data sources and construct an interaction-augmented KG. Based on this graph, we design a two-stage representation learning algorithm for collaboratively learning entity-oriented user preference and preference-enhanced entity representation. 
The obtained entity representation is able to encode implicit entity preference of related users with high-order connectivity on the KG. 
Second, we design a user preference guided discriminator for evaluating the plausibility of a candidate entity given a query.  Besides original KG data, our discriminator is able to utilize the learned preference-enhanced entity representations. Third, we design a query-specific entity generator for producing hard negative entities.
Its major role is to improve the discriminator by learning to sample negative samples from the candidate pool.  

Our approach adopts a ``safer and more careful'' way to utilize user interaction data for the KGC task. We design an elaborate collaborative learning algorithms for learning implicit entity preference of users from their interaction data.  Our generator is relatively isolated from user interaction data, and improves itself according to the feedback from the discriminator.  The discriminator takes entity-oriented user preference as input, and gradually enhances the evaluation capacity in order to defend the increasingly hard fake samples generated by the generator. 
Such an approach is effective to alleviate the issues about data heterogeneity and semantic complexity that were raised earlier. 
To evaluate our approach, we construct extensive experiments  on three real-world datasets. Extensive experiments have demonstrated the effectiveness of our approach on the KGC task, especially for entities with relatively sparse triples.

The rest of this paper is organized as follows. We  first introduce the related work in Section 2. Then, the preliminary and the proposed approach are presented in Section 3 and 4, respectively.  The experimental results are summarized in section 5, and we conclude the paper in section 6.

%% file: sec-rel.tex
\section{Related Work}
Our work is closely related to the studies on knowledge graph completion (KGC), 
collaborative  recommendation and KGC models,  and generative adversarial networks~(GAN).

\paratitle{Knowledge Graph Completion.} 
For the KGC task, various methods have been developed in the literature by adopting different technical approaches.
Translation-based embedding methods, \eg TransE~\cite{Bordes-NIPS-2013} and its variants~\cite{Wang-AAAI-2014,Lin-2015-AAAI}, model relational fact as directed translation from head entity to tail entity. 
Semantic matching based methods~\cite{Nickel-ICML-2011,Yang-CORR-2014,Troillon-ICML-2016,Dettmers-AAAI-2018} serve as another line of research, which try to learn triple plausibility in relational semantic space with bilinear semantic matching.
More recently, Graph Neural Network (GNN)~\cite{Kipf-2017-ICLR,GAT-2018-ICLR} has received much attention  as an effective technique to learn node embeddings over graph-structured data. Several studies try to utilize GNN to capture semantic relations on the KG, such as relational convolution~\cite{Schlichtkrull-2018-ESWC} and structural convolution~\cite{Shang-2019-AAAI}.
 However, these methods mainly focus on modeling KG graph structure, which cannot effectively integrate user interaction data.

\paratitle{Collaborative Recommendation and KGC Models.}
Recently, several studies try to develop collaborative models for the two tasks of item recommendation and KGC, including co-factorization model~\cite{Piao-2018-ESWC}, relation transfer~\cite{Cao-2019-WWW}, multi-task learning~\cite{MKR-WWW-2019} and graph neural networks~\cite{Wang-2019-KDD}.
%
\ignore{Piao et al.~\cite{Piao-2018-ESWC} studied knowledge transfer between the two tasks for the specific domain of items, via a co-factorization model (CoFM). Zhang et al.~\cite{Zhang-SIGIR-2018} proposed a knowledge graph representation learning approach to embed heterogeneous entities.
Wang et al.~\cite{MKR-WWW-2019} associated the two tasks through a cross\&compress units, which automatically share latent features and learn high-order interactions between items in recommender systems and entities in the knowledge graph.
Cao et al.~\cite{Cao-2019-WWW} proposed to transfer the relation information in KG, so as to understand the reasons that a user interacted with an item.}
In these studies,  either shared information is modeled or the same representation space is adopted.
As we discussed, user interaction data is very noisy, and it may be problematic to simply combine the two kinds of data sources. 
Especially, most of these works have set up two optimization objective considering improving both recommendation and KGC. As a comparison, we only consider the KGC task, and user interaction data is only utilized as an auxiliary source. 
Besides, a series of works~\cite{Zhang-KDD-2016,Huang-2018-SIGIR,Sun-2018-RecSys} have been proposed to incorporate knowledge graph to improve the quality and explainability of recommendation.

\ignore{
In general, existing knowledge aware recommender systems can be classified into three categories: (1) Embedding-based methods~\cite{Zhang-KDD-2016,Wang-WWW-2018,Huang-2018-SIGIR} pre-process a KG with knowledge graph embedding algorithms, then incorporate learned entity embeddings into recommendation. (2) Path-based methods~\cite{Yu-WSDM-2014,Hu-KDD-2018,Wang-AAAI-2019} explore various patterns of connections among items in a KG (a.k.a meta-path or meta-graph) to provide additional guidance for recommendations. (3) Hybrid methods~\cite{Wang-2018-CIKM,Sun-2018-RecSys} combine the above two categories and learn user/item embeddings by exploiting the structure of KGs.	
}

\paratitle{Generative Adversarial Networks.} GANs~\cite{GAN-2014-NIPS,gan-survey} have been one of the most breakthrough learning techique in recent years. 
The GAN framework provides a general, effective way to 
estimate generative models via an adversarial process, in which we simultaneously train two models namely generator and discriminator. 
The original GAN~\cite{GAN-2014-NIPS} aims to generate realistic simulation pictures with continuous data representation. Recently, there are quite a few studies that adapt GAN to model data with discrete graph structure, such as graph data~\cite{GraphGAN-2018-AAAI} and heterogenous information network~\cite{HeGAN-KDD-2019}.
These works mainly focus on general graph based tasks (\eg node classification), which  are not directly applicable to our  task.
Especially, GAN has also been used in knowledge graph completion~\cite{KBGAN-2018-NAACL,Wang-2018-AAAI}. 
Their core idea is to enhance the training of existing KGC methods by generating high-quality negative samples, which do not consider other external signals.


Compared with these studies, our focus is to leverage user interaction data for the KGC task with an adversarial learning approach.  We design an elaborate model architecture to effectively fuse user interaction data in the  
discriminator, and utilize a separate generator to produce high-quality ``fake samples'' to help improve the discriminator.  


%% file: sec-pre.tex
\section{Preliminary}
\label{preliminary}

In this section, we first introduce 
the KGC task, then describe the construction details of interaction-augumented knowledge graph based on entity-to-item alignment, and finally present our task.

\paratitle{Knowledge Graph Completion (KGC).} A knowledge graph typically organizes fact information as a set of   triples, denoted by $\mathcal{T}_{KG} = \{\langle h,r,t \rangle|h,t \in \mathcal{E}, r \in \mathcal{R}\}$, where $\mathcal{E}$ and $\mathcal{R}$ denote the entity set and relation set, respectively. A triple $\langle h,r, t \rangle$ describes that there is a relation $r$ between head entity $h$ and tail entity $t$ regarding to some fact. For example, a triple $\langle \textsc{Avatar}, \textsc{directedBy}, \textsc{JamesCameron} \rangle$ describes that the movie of \emph{``Avatar''} is directed by \emph{``James Cameron''}. 
Since not all the facts have corresponding triples in KG,  the KGC task aims to automatically predict triples with missing entities,  either  a tail entity  $\langle h, r, ? \rangle$ or a head entity $\langle ?, r, t \rangle$.
Without loss of generality, in this paper, we only discuss the case with a missing tail entity, \ie $\langle h, r, ? \rangle$. 
For convenience, we call a KG triple with a missing entity \emph{a query}, denoted by $q=\langle h, r, ? \rangle$.
A commonly adopted way by KGC methods is to embed entities and relations into low-dimensional latent space~\cite{Bordes-NIPS-2013,Yang-CORR-2014}, and then develop a scoring function for predicting the plausibility of a triple. Hence, we introduce  
$\bm{v}_h \in \mathbb{R}^K$, $\bm{v}_r  \in \mathbb{R}^K$ and $\bm{v}_t  \in \mathbb{R}^K$ to denote the embeddings for head entities, relations and tail entities, respectively. 


\paratitle{User Interaction.} 
In online systems, we can obtain rich use interaction data with items. 
Formally,  user-item interaction data can be characterized as a set of triples $\mathcal{T}_{UI} =\{\langle u, r_{int}, i \rangle ,|u \in \mathcal{U}, i \in \mathcal{I}\}$, where $\mathcal{U}$ and $\mathcal{I}$ denote the user set and item set respectively, and the triple $\langle u, r_{int}, i \rangle$ indicates that there is an observed interaction $r_{int}$ (\eg purchases and clicks) between user $u$ and item $i$. According to specific tasks or datasets, we can define multiple kinds of user-item interaction relations. Here, for simplicity, we only consider a single interaction relation $r_{int}$.
An interesting observation is that a KG entity usually corresponds to an online item in user-oriented application systems~\cite{Zhao-DI-2019}. For instance, the Freebase movie entity 
``\emph{Avatar}'' (with the Freebase ID \emph{m.0bth54}) has an entry of a movie item in IMDb (with the IMDb ID \emph{tt0499549}). Such a correspondence is called \emph{entity-to-item alignment} across KG and online application systems. 

\paratitle{Interaction-Augmented Knowledge Graph.} Considering the overlap between KG entities and online items, we introduce an extended entity graph to unify the KG information and user interaction data. 
The extended knowledge graph consists of a union set of triples based on KG and online systems:  $\mathcal{G}=\{\langle h,r,t \rangle |h,t \in \mathcal{\widetilde{E}}, r \in \mathcal{\widetilde{R}}\}$, where $\mathcal{\widetilde{E}}=\mathcal{E} \cup \mathcal{U} \cup \mathcal{I}$, $\mathcal{\widetilde{R}} = \mathcal{R} \cup \{r_{int}\}$ and $\mathcal{G} = \mathcal{T}_{KG} \cup \mathcal{T}_{UI}$. 
A major difference with traditional KG is the incorporation of user nodes and user interaction with items into the graph.
We introduce a general placeholder $n$ ($n_j$ and $n_k$) to denote any node on the graph. 
Note that although a KG entity has a corresponding item, we only keep  a single node for a KG entity in the graph. 
Since our task is to leverage user interaction data for learning useful evidence to the KGC task, we organize the entity graph in a user-oriented layer-wise structure. Specially, user nodes are placed on the first layer, then the aligned entities (which correspond to online items) are placed on the second layer. The other nodes are organized in layers according to their shortest distance (\ie minimum hop number) for arriving at any user node.   Let $d_n$
denote the minimum hop number from a node $n$ to user nodes. We can see that $d_n=0, \forall n \in \mathcal{U}$, and $d_n=1, \forall n \in \mathcal{I}$, and $d_n>1, \forall n \in \mathcal{E}\setminus \mathcal{I}$. In this way, entities with the same distance will be placed at the same layer.


\paratitle{Task Description.} Given a query triple $\langle h,r,? \rangle$ or $\langle ?,r,t \rangle$, we aim to predict the missing entity  given both the KG information and user interaction data.
In what follows, we will  focus on the former query case for describing our approach.
While, our experiments will consider both cases for evaluation.  

%% file: sec-model.tex
\section{The proposed approach}

In this section, we present the proposed approach, \emph{UPGAN~(User Preference enhanced GAN)}, for the KGC task by leveraging user interaction data based on  adversarial learning.



\subsection{Overview}
As discussed earlier, user interaction data is quite different from KG data in intrinsic characteristics. It is likely to bring irrelevant information or even noise if  simply integrating it into the KGC method.  
Considering data heterogeneity and semantic complexity, we design an adversarial learning approach to utilizing useful information from user interaction data for the KGC task. 


\ignore{For this first issue, 
we propose a collaborative representation learning  over the interaction-augmented KG. It divides the learning process into two stages: the first stage learns entity-oriented user preference by propagate entity semantics to users, and the second learns performance-enhanced entity representations by collection the evidence from users to the target entity. 
}




We set up two components with different purposes for the KGC task, namely prediction component~(\ie generator $G$) and evaluation component~(\ie discriminator $D$). 
The generator $G$ produces a candidate answer for the missing entity, and the discriminator $D$ evaluates the plausibility of the generated answer by $G$.
The two components force each other to improve in a mutual reinforcement way. 
Our  focus is to train a capable discriminator that is able to leverage KG information for the KGC task, and the role of the generator is to improve the discriminator and help the fusion of user interaction data. 
In this way, we can fully utilize useful evidence from user interaction data in the discriminator, and meanwhile avoid  direct influence of user interaction data on the KG semantic space modeled by the generator.  

Following  GANs~\cite{GAN-2014-NIPS,KBGAN-2018-NAACL,GraphGAN-2018-AAAI}, we cast our problem as  a minimax game between two players, namely generator $G$ (parameterized by $\theta^G$) and discriminator $D$ (parameterized by $\theta^D$), for our KGC task:
\begin{equation}
\begin{split}
\min \limits_{\theta^G} \max \limits_{\theta^D} \quad &\quad \quad \mathbb{E}_{\langle h,r,t \rangle \sim P_{\mathcal{T}_{KG}}} \log D(t | h, r;\mathcal{G}, \theta^D) \\
\quad &+ \quad \mathbb{E}_{\langle h,r,? \rangle \sim P_{\mathcal{T}_{KG}}, a \sim G} \log\big(1-D(a | h, r; \mathcal{G}, \theta^D
)\big).
\end{split}
\end{equation} 
where $a\sim G(h,r; \theta^G)$ denotes a generated entity by the generator. 
The discriminator would drive the generator to produce more better candidates, and the generator would improve the discriminator by providing more hard fake samples. By repeating such a mutual improvement process, we expect a more effective KGC solution can be derived. 
Note $\mathcal{G}=\mathcal{T}_{KG} \cup \mathcal{T}_{UI}$, consisting of KG triples and user-item interaction triples, has been incorporated into the discriminator $D$.
To model the information on the heterogeneous graph $\mathcal{G}$, we develop a collaborative representation learning algorithm based on graph neural networks for extracting useful user preference information from user interaction data. 

\begin{figure}[!tb]
	\small
	\centering 
	\includegraphics[width=0.43\textwidth]{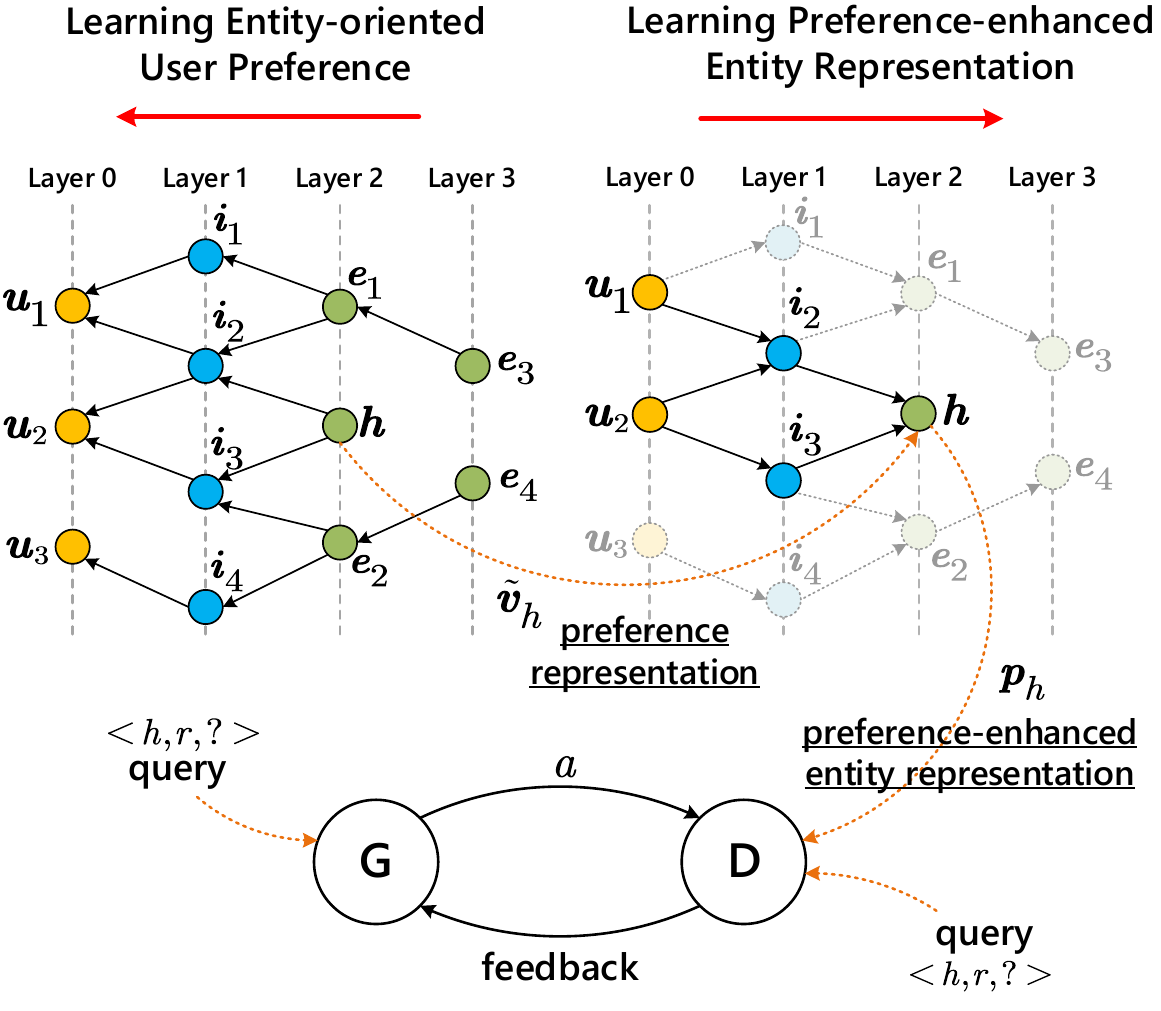}
	\caption{The overview of the proposed UPGAN model. The orange, blue and green nodes represent the users, items interacted with users, and entities in KG, respectively.} 
	\label{fig-model} 
\end{figure}

We present an overall sketch of the proposed approach in Fig.~\ref{fig-model}.
In what follows, we first introduce how to learn suitable representations from $\mathcal{G}$, and then describe the discriminator and generator.

\subsection{Collaborative Representation Learning over Interaction Augmented KG}
\label{pre-learning}
As shown in Fig.~\ref{fig-model}, user interaction data explicitly reflects user preference at the item level, and we would like to learn  and utilize implicit entity-oriented preference of users in the semantic space of KG.
Our solution is to learn effective node embeddings over the interaction-augmented KG $\mathcal{G}$, which is expected to encode useful preference evidence for enhancing KG entity representations. 
A straightforward method is to treat all the graph nodes equally and employ a standard graph neural network model to learn node embeddings.
However, 
it may incorporate irrelevant information or noise into node representations due to node heterogeneity.  
To address this issue, we design an elaborative  two-stage collabarative learning algorithm based on user-oriented graph neural networks. 

\ignore{A key issue  is how to learn useful evidence from user interaction data for the KGC task. User interaction data explicitly reflects user preference at the item level, and we need to learn implicit entity-level preference information related to KG semantics. 
As we introduced before, we have constructed a user-oriented knowledge graph for unifying KG and user interaction data. We consider utilize this graph to map explicit user-item interaction into implicit user-entity relatedness. 
A straightforward method is to treat all the nodes in this graph equally the same and employ a standard graph neural network model to learn node embeddings.
However, large-scale user interaction data is likely to contain noise affecting the final semantic representation. 
To address this problem, we design an elaborative  two-stage learning algorithm based on user-oriented graph neural networks. 
}

\subsubsection{Learning Entity-oriented User Preference} 
\label{entity-oriented}
Recall that user nodes are placed at the bottom layer, and other entity nodes are at a higher layer. In the first stage, we preform the information propagation from KG entities to users.  
The update strategy is a combination between the original embedding and the received embeddings from forward triples:
\begin{equation}
\label{eq-user-learn}
\bm{\tilde{v}}_{n_j} = \sigma(\bm{W}_0^D\bm{v}_{n_j}+ \sum_{\langle n_j, r, n_k \rangle \in \mathcal{F}_{n_j}} \frac{1}{|\mathcal{F}_{n_j}|}\bm{W}_r^D\bm{\tilde{v}}_{n_k}).
\end{equation}
where $\bm{v}_{n_j} $ is the original learned or initialized node representation,  $n_j/n_k$ denotes a node on the graph (can be a user, item or entity),  $\mathcal{F}_{n_j} = \{\langle n_j, r, n_k \rangle | d_{n_j} = d_{n_k}-1, \langle n_j, r, n_k \rangle \in \mathcal{G}\}$ denotes the set of forward triples (an entity links to another connected entity at the next layer) for entity $n_j$, and $\bm{W}_0^D$ and $\bm{W}_r^D$ denote the transformation matrices for the original representation and relation $r$, respectively. 
With this update formula, a node on the graph can collect related entity semantics from its upstream neighbors. By organizing nodes in layers,  the entities closer to users have a greater impact on user preference.
The update for user embeddings is performed at the last step, which alleviates the influence of noisy interaction data. 
Another merit is that the propagation implicitly encodes path semantics into the node representations, which has been shown important to consider in the KGC task~\cite{Guu-2015-EMNLP,PTransE-EMNLP-2015}. 
When this stage ends, each user node will be learned with a preference representation $\tilde{\bm{v}}_u$ based on Eq.~\ref{eq-user-learn}, encoding her/his preference over entity-level semantics.

\subsubsection{Learning Preference-enhanced Entity Representation} In the second stage,  given a query triple $\langle h, r, ? \rangle$, we would like to collect user preference information over entity semantics on the graph regarding to the target entity $h$. For example, in Fig.~\ref{fig-example}(b), knowing the preference of user ``\emph{Alice}'' is helpful to answer the query regarding to the director for entity ``\emph{Avatar}''.
For this purpose, we perform an inverse aggregation from user nodes to the target entity as follows: 
\begin{equation}
\begin{split}
\label{eq-user-select}
\bm{p}_{n_j} &= \sum_{\langle n_j, r, n_k \rangle \in \mathcal{B}_{n_j}}\alpha_{n_j, r, n_k}\bm{p}_{n_k}, 
\end{split}
\end{equation}
where $\mathcal{B}_{n_j} = \{\langle n_j, r, n_k \rangle |d_{n_j} = d_{n_k}+1, \langle n_j, r, n_k \rangle \in \mathcal{G}\}$ denotes the set of backward triples  (an entity links to another connected entity at the previous layer)  for entity $n_j$, and $\alpha_{n_j, r, n_k}$ is the  attention coefficient for aggregation defined as 
\begin{equation}
\begin{split}
\alpha_{n_j,r,n_k}&=\frac{\exp(\pi(n_j,r,n_k))}{\sum_{\langle n_j, r', n_{k'} \rangle \in \mathcal{B}_{n_j}}\exp(\pi(n_j,r',n_{k'}))}, \\
\pi(n_j,r,n_k)&=\text{LeakyReLU}(\bm{w}^\top[\bm{W}_r^D\bm{\tilde{v}}_{n_j};\bm{W}_r^D\bm{p}_{n_k}]). 
\end{split}
\end{equation}
where $\bm{\tilde{v}}_{n_j}$ is the learned representation in Section~\ref{entity-oriented}. Before running the 
aggregation procedure, we first initialize $\bm{p}_{n}$ as $\bm{\tilde{v}}_{n}$.
Given a target entity,  our aggregation update indeed spans a tree-like structure (See Fig.~\ref{fig-model}), and only reachable nodes on the graph are activated in this process.
When this stage ends, we can derive an updated representation for the target entity $h$, which encodes the preference information passed from activated user nodes, denoted by $\bm{p}_h$.

\subsubsection{Discussion} 
We have designed an elaborate two-stage learning algorithm over the interaction-augmented KG. The update in both stages is directed. The first stage propagates entity semantics to user nodes, which aims to learn entity-oriented user preference; the second stage collects the learned user preference at the target entity, which aims to learn preference-enhanced entity representations.  
When the involved weight parameters are fixed, it can be proved that $\bm{p}_h$ is indeed a linear combination of user preference representations (learned in the first stage), given the fact that we aggregate the information by layer and start from the first layer of user nodes. It can be formally given as:
\begin{equation}
\label{eq-user-preference}
\bm{p}_h = \sum_{u \in \mathcal{U} } w_{h,u} \tilde{\bm{v}}_u,
\end{equation}
where $\tilde{\bm{v}}_u$ is the user embeddings learned in the first stage (Eq.~\ref{eq-user-learn}), and $w_{h,u}$  (set to zero for unactivated users) can be computed according to the accumulative attention coefficients along the paths from user $u$ to  target entity $h$. 
Indeed, these activated users are high-order connectable nodes to the target entity. 
Besides the learned semantic representation $\bm{v}_h$, we enhance the entity representation using the entity-level preference of the users with high-order connectivity.

\subsection{User Preference Guided Discriminator}\label{sec-user-pref}
In our approach, the major function of the discriminator is to distinguish between real and fake answers given the query.
Compared with previous GAN-based KGC methods~\cite{KBGAN-2018-NAACL,Wang-2018-AAAI},
a major difference is that we would incorporate the learned preference-enhanced entity representations for improving the discriminator. 


\subsubsection{Discriminator Formulation.} 
Our discriminator $D(t | h,r;\mathcal{G}, \theta^D)$ evaluates whether the entity $t$ can be the answer to a given query $\langle h, r, ? \rangle$ by computing the following probability: 
\begin{equation}
	\label{prob}
	\centering
	\begin{split}
	D(t|h,r; \mathcal{G}, \theta^D) &= \frac{1}{1 + \exp (-s(h,r,t; \mathcal{G}, \theta^D))},
	\end{split}
\end{equation}
where $s(\cdot)$ is the score function measuring the plausibility of the triple $\langle h, r, t \rangle$. Here, we give a general form for $s(\cdot)$, and many previous methods can be used to instantiate it, such as TransE~\cite{Bordes-NIPS-2013} and DistMult~\cite{Yang-CORR-2014}. 
We incorporate the preference-enhanced entity representation $\bm{p}_h$ for improving the evaluation capacity of the discriminator as follows: 
\begin{equation}
\label{eq-score-D}
	s(h,r,t ; \mathcal{G}, \theta^D) =(\bm{W}_2 \bm{v}_t + \bm{b}_2)^{\top} \cdot \tanh(\bm{W}_1 \bm{x}_q + \bm{b}_1),
\end{equation}
where $\bm{W}_1, \bm{W}_2$ and $\bm{b}_1, \bm{b}_2$ are parameter matrices or vectors, $s(\cdot)$ takes as input the query embedding $\bm{x}_q$ and candidate entity embedding $\bm{v}_t$, and $\tanh(\cdot)$ is incorporated as a non-linear transformation function that can be replaced by other functions.  $\bm{x}_q$  is composed of two parts: the learned entity embeddings using KG information and the enhanced entity representation from user interaction data,  formally given as
\begin{equation}\label{eq-p_h_xq}
	\bm{x}_q = [\underbrace{\bm{v}_h \odot \bm{v}_r}_{\text{KG information} }; \underbrace{\bm{p}_h \odot \bm{v}_r}_{\text{preference information}}],
\end{equation}
where $\bm{p}_h$ is defined in Eq.~\ref{eq-user-preference} reflecting the related user preference regarding to entity $h$. In this way, user preference over KG entities on the graph $\mathcal{G}$ has been considered into the discriminator.
 A good candidate answer should not only match the query well in the KG, but also meet the semantic requirement of the entity  preference of users with high-order connectivity.

\ignore{
\begin{equation}
	\centering
	\begin{split}
	s(h,r,t ; \mathcal{G}, \theta^D) &=\bm{v}_t^{\top} \cdot \tanh (\bm{W}_1 \bm{x}_q + \bm{b}_1), \\
	\bm{x}_q &= [\underbrace{\bm{v}_h \odot \bm{v}_r}_{\text{KG information} }; \underbrace{\bm{p}_h \odot \bm{v}_r}_{\text{preference information}}],
	\end{split}
\end{equation}
where $s(\cdot)$ takes as input the query embedding $\bm{x}_q$ and candidate entity embedding $\bm{v}_t$, and $\bm{x}_q$ is composed of two parts: the learned entity embeddings using KG information and the enhanced entity representation from user interaction data. Here, $\bm{p}_h$ is defined in Eq.~\ref{} reflecting the related user preference regarding to entity $h$. In this way, user preference over KG entities has been considered into the discriminator. A good candidate answer should not only match the query well in the KG, but also meet the semantic requirement of the entity  preference of users with high-order connectivity.
}

\subsubsection{Discriminator Loss} To optimize the discriminator, we consider two cases for computing the loss. 
First, the real answer entity $t$ to the query $\langle h, r, ? \rangle$ on the knowledge graph $\mathcal{T}_{KG}$ should be recognized as positive by the discriminator. Second, the discriminator tries to identify the generated answer by the generator $G(h,r;\theta^G)$ as negative. The loss of the two cases can be given as follows:
\begin{equation}
\label{eq-loss-D}
\begin{split}
	\mathcal{L}^D = &\mathbb{E}_{\langle h,r,t \rangle \sim P_{\mathcal{T}_{KG}}} - \log D(t|h,r)+
	\\
	&\mathbb{E}_{\langle h,r,? \rangle \sim P_{\mathcal{T}_{KG}}, t' \sim G(h,r;\theta^G)} - \log (1 - D(t' |h,r))+ \lambda^D || \theta^D ||^2_2,
\end{split}
\end{equation}
where $\lambda^D > 0$ controls the regularization term to avoid overfitting. Given a query, the real answers from the KG population is considered as the positive cases, and the generated entities
from the generator $G$ as the negative cases.
The parameter $\theta^D$ of the discriminator can be optimized by minimizing $\mathcal{L}^D$.
 Note that although we describe the learning of $\bm{p}_h$ and the discriminator in different sections, they are bound through the discriminator objective and will be learned jointly. 
 With increasingly hard samples from the generator, the discriminator jointly optimizes its own parameters and the involved parameters in Section 4.2. In this way, the entity-oriented user preference $\bm{\tilde{v}}_u$ and enhanced entity representation $\bm{p}_h$ are gradually transformed into a suitable representation for the KGC task.

\subsection{Query-specific Entity Generator}
In our approach, the major function of the generator is to provide high-quality negative entities to improve the discriminator.  We design a query-specific entity generator by sampling from the candidate entity pool.  Since user interaction data itself is likely to contain noise, the generator would not utilize any user interaction data and  model a pure KG semantic space. 

\subsubsection{Generator Formulation} For each query $\langle h, r, ?\rangle$, we assume that a candidate entity set  $\mathcal{C}^{h,r} \subset \mathcal{E}$ can be first constructed, \eg using existing KGC methods or random sampling.  Then, our generator defines a distribution over the candidate set and samples from it. Given a query $q=\langle h, r, ? \rangle$, we compute the query representation $\bm{v}_q^G$ as
\begin{equation}
\bm{v}_q^G = \bm{v}_h \odot \bm{v}_r.
\end{equation}
We can  implement $\bm{v}_q^G $ in other ways as needed. 
Note that KG embeddings $\bm{v}_h$ and $\bm{v}_r$ are not necessarily the same as those in the discriminator. To enhance the robustness of our generator, we concatenate the query representation with a noise $\bm{z}\sim \mathcal{N}(\bm{0}, \sigma^2 \bm{I})$, which is a Gaussian distribution with zero mean  and covariance $\sigma^2 \bm{I}$:
\begin{equation}
\bm{e}_q^G = [\bm{v}_q^G; \bm{z}].
\end{equation}
Finally, the concatenated vector is fed into a Multi-Layer Perceptron~(MLP), which is activated with non-linear function LeakyReLU. The probability distribution to sample a candidate entity from  $\mathcal{C}^{h,r}$ is defined as:
\begin{equation}
\label{generator-probability}
G(a|h, r;\theta^G)= \frac{\exp(\text{MLP}(\bm{e}_q^G)\cdot \bm{v}_a)}{\sum_{t'  \in \mathcal{C}^{h,r} } \exp(\text{MLP}(\bm{e}_q^G)\cdot \bm{v}_{t'})}.
\end{equation}
With this  distribution, we sample  $n_G$ entities from the candidate set, which are taken as input for the discriminator as negative samples. 

\subsubsection{Policy Gradient.}  Since sampling an entity from the candidate set is a discrete process, we do not directly optimize the loss for the generator.  Here, we follow KBGAN~\cite{KBGAN-2018-NAACL} to adopt  policy gradient~\cite{PolicyGradient-1999-NIPS} for parameter learning. A key point is how to set the reward function appropriately.  Here, we utilize the feedback of the discriminator as the reward signal to guide the learning of the generator. 
\begin{equation}
\label{eq-reward-G}
R(a|h,r) = \frac{\exp(s(a|h,r;\mathcal{G},\theta^D)) }{\sum_{t' \in \mathcal{C}^{h,r}} \exp(s(t'|h,r;\mathcal{G},\theta^D)) } - b,
\end{equation}
where  the score function $s(\cdot)$ is defined in Eq.~\ref{eq-score-D} and we set $b=\frac{1}{|\mathcal{C}^{h,r}|}$ as the bias. Here, we incorporate the bias $b$ by considering uniform sampling as a reference. 
When a sample receives a larger probability by the discriminator than  the  average, it would  be assigned with a positive reward by our approach. 
Formally, we optimize the following loss for the generator:
\begin{equation}\label{g-loss}
\mathcal{L}^G = \mathbb{E}_{\langle h,r,? \rangle \sim P_{\mathcal{T}_{KG}}, a \sim G(h,r;\theta^G)} R(a|h,r) + \lambda^G ||\theta^G||^2_2,\\
\end{equation}
where $\lambda^G > 0$ controls the regularization term to avoid overfitting. To optimize the above loss, the policy used by the generator would punish the trivial negative entities by lowering down their corresponding probability, and encourage the network to assign a larger probability to the entities that can bring higher reward.

\subsection{Optimization and Discussion}
In this part, we discuss the model optimization and comparison with previous works.

To learn our model, we first pretrain the discriminator component with  training data. 
Then, we  follow the standard training algorithms for GAN-based models~\cite{GAN-2014-NIPS} by alternating between the $G$-step and $D$-step at an iteration. 
We adopt a mini-batch update strategy. For each training triple $\langle h, r, t \rangle$ in a batch, the generator will first randomly sample $n_C$ entities from the entire entity set (excluding observed true answers) as the candidate pool $\mathcal{C}^{h,r}$. Since the entire entity set is likely to contain false negatives (\ie true answers),  we empirically find that  $n_C$  should not be set to a very large number. After that, the generator $G$ samples $n_G$ entities from the $n_C$ candidates as negative samples using Eq.~\ref{generator-probability}. 
Then, we update the parameters of $G$ according to the loss in Eq.~\ref{g-loss} using policy gradient~\cite{PolicyGradient-1999-NIPS}. 
For the discriminator, given a query from the training set, it minimizes the loss in Eq.~\ref{eq-loss-D} over the real answer and the $n_G$ fake samples from the generator.

\ignore{\textcolor{blue}{UPGAN is trained with minibatches. After sampling a batch of triples, we first find users that will be activated in this batch and perform a sub-graph entity-to-user information propagation according to Section 4.2.1. Then we learn the preference-enhanced entity representations only for the query entities in the sampled batch according to the method in Section 4.2.2. Since we span a tree-like structure for this procedure, it can be efficiently implemented with tree traverse algorithms. The discriminator is trained with positive cases sampled from $\mathcal{T}_{KG}$ and negative cases generated by G. At the same time, we can compute rewards for G with scoring of disciminator. And G is optimized with policy gradient~\cite{PolicyGradient-1999-NIPS}. 
}
}

Note that the parameters involved in the graph neural networks in Section 4.2  will be also optimized in the learning process of $D$, since $D$ directly uses the learned node embeddings from it. 
We first identify the users that are activated (\ie reachable) by the entities from  a batch. 
After that, we employ these users as seeds to construct a subgraph for local parameter update. 
Based on the subgraph, we perform an entity-to-user information propagation according to Section 4.2.1, and then learn the preference-enhanced entity representations only for the query entities in the sampled batch according to Section 4.2.2. Since we span a tree-like structure for this procedure, it can be efficiently implemented with tree traverse algorithms.
To encourage the discriminator to estimate soft probabilities, we adopt the label smoothing trick~\cite{Salimans-NIPS-2016}  to train our UPGAN. 
Although there have been a few studies which either adopt GAN or utilize user interaction data for improving the KGC task, our approach has two major differences. 
First, our adversarial approach is developed based on an effective two-stage learning algorithm for integrating both entity semantics and user preference. As a comparison,  user interaction information has been seldom considered in previous GAN based methods for the KGC task. 
Second, we do not directly incorporate the learned information  from user interaction into the generator.
Its major role is to improve  the discriminator by producing high-quality fake samples.
To our knowledge, it is the first time that user interaction data has been utilized for the KGC task in an adversarial learning approach. 

\ignore{
First, we don't seek to learn shared entity representations in the same space for two data resources. 
Instead, our discriminator only utilizes the learned user preference to enhance the evaluation capacity for a candidate entity, while our generator adopts a more loose incorporation of user preference information by sampling it from Gaussian distribution.  This strategy makes our approach is able to fully utilize the useful information from user interaction data, and meanwhile keep the prediction component more resistible to noise. Second, our adversarial framework is developed based on an effective two-stage learning algorithm for integrating both entity semantics and user preference. While, such user interaction information has been seldom considered in graph based GANs. The learned user preference plays an important role in integrating the discriminator and generator. 
}

%% file: sec-exp.tex
\section{experiment}
In this section, we perform the evaluation experiments for our approach on the KGC task. We first introduce the  experimental setup, and then report the results and detailed analysis.

\ignore{We aim to answer
the following research questions:
\begin{itemize}
	\item \paratitle{RQ1}: How does XXGAN perform compared with state-of-the-art knowledge graph embedding methods?
	\item \paratitle{RQ2}: When the interaction data should be used?
	\item \paratitle{RQ3}: Why the interaction data is useful in the KGC solution?
\end{itemize}
}

\subsection{Dataset Construction}

In our setting, we need an aligned linkage between KG data and user interaction data. Here, we adopt the KB4Rec  dataset~\cite{Zhao-DI-2019}  to construct the evaluation datasets, containing the alignment records between Freebase entities~\cite{freebase} and online items from three domains.

  Freebase stores facts by triples of the form $\langle$\emph{head}, \emph{relation}, \emph{tail}$\rangle$, and we use 
 the last public version released on March 2015.
 The three user interaction datasets are  \textsc{MovieLens} movie~\cite{Harper2016The}, \textsc{LFM-1b} music~\cite{Schedl2016The} and \textsc{Amazon} book~\cite{He2016Ups}.
For all datasets, we only keep the interactions related to the linked items.
The \emph{LFM-1b} music dataset is very large, and we take the subset from year 2012; while for the \emph{MovieLens} 20m dataset, we take the subset from year 2005 to 2015.
Following~\cite{Rendle-2010-WWW}, we only keep the $k$-core dataset, and filter out unpopular items and inactive users with fewer than $k$ interaction records, which is set to 10 for the music dataset and 5 for the other two datasets.

After preprocessing the three user interaction datasets, we take the left aligned entities as seeds, and generate the KG subgraph by 
performing breadth-first-search in each domain. We aim to examine the performance improvement of queries about  both aligned entities and their reachable entities via a few hops on the KG. In our experiments, we set the maximum BFS hop to be four. Following~\cite{Bordes-NIPS-2013,Toutanova-EMNLP-2015}, we removed relations like \emph{<book.author.works\underline{~}written>} which just reverses the head and tail compared to the relations \emph{<book.written\underline{~}work.author>}.
We also removed relations that end up with non-freebase string, \eg like \emph{<film.film.rottentomatoes\underline{~}id>}. 
To ensure the KG quality, we filter infrequent entities with fewer than $k$ KG triples, which is set to 3 for the book dataset and 10 for the other two datasets. We summarize the statistics of three datasets after preprocessing in Table~\ref{tab-data}. Overall, the user interaction data in the book domain is sparser than the other two domains.
Furthermore, for each domain, we randomly split it into training set, validation set and test set with a ratio of 8:1:1. 


\begin{table}[htbp]
  \centering
  \caption{Statistics of our datasets after preprocessing.}
  \label{tab-data}%
    \begin{tabular}{|c|l|r|r|r|}
		\hline
		\multicolumn{2}{|c|}{Dataset}& Movie & Music & Book\\
		\hline
		\hline
		\multirow{3}{*}{\tabincell{c}{User\\Interaction}}&\#Users 	&	61,859&	57,976&	75,639\\
		&\#Items 	&	17,568&	55,431&	22,072\\
		&\#Interactions 	&	9,908,778&	2,605,262&	831,130\\
		\hline
		\multirow{3}{*}{\tabincell{c}{Knowledge\\Graph}}&\#Entities	&56,789	&108,930	&79,682	\\
		&\#Relations	&47	&45	&38	\\
		&\#Triplets 	&953,598	&914,842	&400,787	\\
		\hline
    \end{tabular}%
\end{table}%

\subsection{Experimental Setting}
This part presents the basic experimental settings. 
\subsubsection{Evaluation Protocol} We follow~\cite{Bordes-NIPS-2013} to cast the KGC task as a ranking task for evaluation. 
For each test triple $\langle h,r,t \rangle$ in a dataset, two queries, $\langle h,r,? \rangle$ and  $\langle?, r, t \rangle$, were issued in the following way. Each missing entity (\ie ground truth) will be combined with the rest entities as a candidate pool (excluding other valid entities). Given a query, a method is required to rank the order of the entities in the candidate list, and a good method tends to rank the correct entity in top positions. 
To evaluate the performance, we adopt a variety of evaluation metrics widely used in previous works, the Mean Rank (MR)~\cite{Bordes-NIPS-2013}, top-$k$ hit ratio (H@$k$)~\cite{Bordes-NIPS-2013}, and Mean Reciprocal Rank (MRR)~\cite{Yang-CORR-2014}. Specifically, MR refers to the average rank of all testing cases, H@$k$ is defined as the percentage of the testing triples that have a rank value no greater than $k$, and MRR is the average of the multiplicative inverse of the rank value for all testing triples.
For all the comparison methods, we learn the models using the training set, and optimize the parameters using the validation set and compare their performance on the test set.


\subsubsection{Methods to Compare} We consider the following methods for performance comparison:
\begin{itemize}
	\item \textbf{TransE}~\cite{Bordes-NIPS-2013}: TransE model introduces translation-based embedding, modeling relations as the translations operating on entities.
	\item \textbf{DistMult}~\cite{Yang-CORR-2014}: It is based on the bilinear model where each relation is represented by a diagonal rather than a full matrix.
	\item \textbf{ConvE}~\cite{Dettmers-AAAI-2018}: It is a link prediction model that uses 2D convolution over embeddings and multiple layers of non-linear features.
	\item \textbf{ConvTransE}:~\cite{Shang-2019-AAAI}: ConvTransE enable the state-of-the-art ConvE to be translational between entities and relations while keeps the same link prediction performance as ConvE.
	\item \textbf{KBGAN}~\cite{KBGAN-2018-NAACL}: It utilizes pretrained KG embedding models as generator to selectively generate hard negative samples, and improves the performances of target embedding models.
	\item \textbf{R-GCN}~\cite{Schlichtkrull-2018-ESWC}:  It is related to a recent class of neural networks operating on graphs, and is developed specifically to handle the highly multi-relational data characteristic of realistic KGs.
	\item \textbf{KTUP}~\cite{Cao-2019-WWW}: It jointly solve recommendation and KGC tasks, transfering the relation information in KG, so as to understand the reasons that a user likes an item.
	\item \textbf{CoFM}~\cite{Piao-2018-ESWC}: It is a multi-task co-factorization model which optimizes both item recommendation and KGC task jointly.
	\item \textbf{KGAT}~\cite{Wang-2019-KDD}: Built upon the graph neural network framework, KGAT explicitly models the high-order relations in collaborative knowledge graph  with item side information.
	\item \textbf{UPGAN}: It is our approach.
\end{itemize}

Our baselines have a comprehensive coverage of the related models. To summarize, we categorize the baselines into several groups shown in Table~\ref{tab:baselines}, according to the \emph{technical approaches} and \emph{utilization of user interaction data}. 
All the models have some parameters to tune. We either follow the reported optimal parameters or optimize each model separately using  validation set.
Following~\cite{Dettmers-AAAI-2018,Shang-2019-AAAI}, we equip semantic-matching based methods with $1-N$ scoring strategy, including DistMult that previously adopted a simple binary entropy cross loss. 

\begin{table}[htbp]
	\centering
	\caption{The categorization of the comparison methods. ``UI'' is the abbreviation for user interaction. }
	\label{tab:baselines}%
	\begin{small}
		\begin{tabular}{|c|c|c|c|}
			\hline
			Category & Translation & Semantic match & GNN\\
			\hline
			\hline
			\multirow{2}{*}{\tabincell{c}{KG}} &\multirow{2}{*}{\tabincell{c}{TransE}} & \multirow{2}{*}{\tabincell{c}{DistMult,ConvE,\\ConvTransE}} & \multirow{2}{*}{\tabincell{c}{R-GCN}} \\
			& & & \\
			\hline
			KG+GAN&\multicolumn{3}{c|}{KBGAN}\\
			\hline
			\tabincell{c}{KG+UI} &\tabincell{c}{KTUP,CoFM} & --- & \tabincell{c}{KGAT} \\
			\hline
			KG+UI+GAN&\multicolumn{3}{c|}{UPGAN (our approach)}\\
			\hline
		\end{tabular}
	\end{small}
\end{table}

\begin{table*}[h]
	\centering
	\caption{Performance comparison of different methods for KGC task on three datasets. We use bold and underline fonts to denote the best and second best performance in each metric respectively. Besides MR, the results are given in precent (\%). }
	\label{tab:results-all}%
	\begin{tabular}{|l | c c c c c| c c c c c| c c c c c|}
		\hline
		\multirow{2}{*}{Models}&\multicolumn{5}{c|}{Movie}&\multicolumn{5}{c|}{Music}&\multicolumn{5}{c|}{Book}\\
		\cline{2-16}
		&MR & MRR& H@1&	H@3&	H@10&	MR& MRR& H@1&	H@3&	H@10&	MR & MRR& H@1&	H@3&	H@10\\
		\hline
		\hline
		TransE 	&	1941&	18.7&	12.3&	20.5&	32.2&	864&	61.7&	53.7&	66.6&	76.9&	5694&	31.7&	25.3&	34.9&	44.1\\
		DistMult&	\textbf{1218}&	\underline{25.2}&	18.4&	\underline{27.8}&	\underline{38.5}&	2153&	68.4&	62.3&	72.5&	79.3&	6676&	\underline{34.9}&	\underline{29.3}&	\underline{37.8}&	45.7\\
		ConvE 	&	1671&	24.6&	18.3&	27.0&	36.9&	1620&	69.3&	63.7&	73.0&	79.4&	4858&	33.0&	27.0&	36.0&	44.3\\
		ConvTransE&	1450&	25.0&	\underline{18.5}&	27.5&	37.8&	1203&	\underline{69.9}&	\underline{63.9}&	\underline{73.8}&	\underline{80.6}&	3995&	33.4&	27.0&	36.8&	45.4\\
		R-GCN 	&	\underline{1261}&	24.4&	18.0&	26.6&	37.0&	1565&	68.4&	62.6&	72.0&	78.9&	6438&	32.8&	27.6&	35.2&	42.1\\
		KBGAN 	&	2324&	20.9&	14.8&	23.2&	33.3&	995&	63.2&	55.8&	67.7&	77.1&	6539&	32.3&	26.2&	35.3&	44.4\\
		\hline
		CoFM&	1936&	18.8&	12.3&	20.6&	32.2&	2204&	62.4&	54.5&	67.1&	77.4&	5695&	31.7&	25.3&	35.0&	44.1\\
		KTUP&	1960&	19.3&	12.7&	21.2&	32.8&	\underline{851}&	62.0&	54.1&	66.8&	77.0&	5456&	32.1&	25.7&	35.3&	44.5\\
		KGAT &	1347&	20.1&	13.8&	22.2&	32.3&	\textbf{593}&	62.5&	53.6&	68.2&	78.4&	\textbf{2670}&	34.1&	27.6&	37.1&	\underline{46.0}\\
		\hline
		UPGAN & 1666&	\textbf{25.9}&	\textbf{18.8}&	\textbf{28.9}&	\textbf{39.4}& 1050& \textbf{71.8}& \textbf{65.8}& \textbf{75.9}& \textbf{82.1}& \underline{3463}& \textbf{37.0}& \textbf{30.6}& \textbf{40.5}& \textbf{48.8}\\
		\hline
	\end{tabular}%
\end{table*}%

\subsubsection{Implementation Details}
For our approach, 
we adopt the DistMult~\cite{Yang-CORR-2014} model to initialize the KG related parameters, and train each individual component to converge for (at most) 1000 epochs. To avoid overfitting, we adopt early stopping by evaluating MRR on the validation set every 20 epochs. We optimize all models with Adam optimizer, where the batch size is set to 4096. The coefficient of L2 normalization is set to $10^{-5}$, and the embedding size is set to 100, and the learning rate is tuned amongst \{0.01, 0.005, 0.001, 0.0005, 0.0001\}. 
The entity embeddings are constrained to have a length no smaller than 1.
In each iteration, we set $n_G$ as 200 and $n_C$ as 1024.
For the generator,  the MLP components contain two hidden layers with the LeakyReLU activation function.


\subsection{Results and Analysis}


 The results of different methods for knowledge graph completion task are presented in Table~\ref{tab:results-all}. It can be observed that:

(1) Among baselines which only use KG data, TransE performs worst since it usually adopts very simple distance function for fitting training triples. 
Three semantic match based methods DistMult, ConvE and ConvTransE give better results than TransE, which have used a more powerful match function for modeling the semantics of a triple. 
The GNN based method R-GCN shows a more competitive performance than TransE, while it performs worse than  semantic match based methods.
Overall,  DistMult and ConvTransE are the best baseline methods. 
\ignore{
For the two GNN based methods,  LENA seems to perform better than R-GCN. A possible reason is that LENA focuses on the modeling of neighborhood information on the KG, and is effective to collect sufficient semantic evidence to perform more reliable reasoning over unknown KG triples.	
}

(2) KBGAN is the only GAN based baseline, which mainly aims to produce high-quality negative samples than random sampling. 
As we can see that, it substantially improves over TransE on all datasets, which indicates the usefulness of adversarial learning.
However, KBGAN only utilizes the information from the KG triples, and its improvement is relatively limited, and cannot perform better than the competitive baselines DistMult and ConvTransE. 
Besides, for a query, DistMult and ConvTransE  adopt a new $1-N$ scoring function~\cite{Dettmers-AAAI-2018} as the enhanced loss by iterating over all the candidate entities. We speculate that the usefulness of $1-N$ scoring strategy is mainly due to \emph{candidate exposure} by  simply treating all the entities from the entire candidate set to be negative.  

(3) Overall, the three methods that jointly utilize KG data and user interaction data seem to give slightly better results than TransE. 
Among these methods, CoFM and KTUP are indeed constructed based on  translation based methods. 
KGAT has developed a collaborative graph neural network for learning the embeddings over the heterogeneous nodes. 
It achieves a better performance on book dataset than the other two datasets. 

(4) Finally, we compare the proposed approach UPGAN with the baseline methods.
 It is clear to see that UPGAN is consistently better than these baselines by a large margin. 
 As shown in Table~\ref{tab:baselines}, our method jointly utilizes the KG and user interaction data using a GAN-based approach.
 Different from the above joint models, we  optimize the performance of the KGC task as the only objective. Especially, we adopt an elaborative way to incorporate the learned user preference. We only utilize the user interaction data in the discriminator, while  the major role of the generator models is to improve the discriminator. The  generator is improved according to the feedback of the discriminator, which can be considered as indirect signal from user interaction data. 

\subsection{Detailed Analysis of Performance Improvement}\label{sec-improvement}
As shown in Table~\ref{tab:results-all}, our proposed approach UPGAN shows a better overall  performance than the  baselines. Here, we zoom into the results and check whether UPGAN is indeed better than baselines in specific cases. 
For ease of visualization, we only incorporate the results of DistMult and ConvTransE as the reference, since they perform generally well among all the baselines.  

\begin{table}[htbp]
	\centering
	\caption{Performance (H@3 in precent) comparison \emph{w.r.t.} different sparsity levels. $\%Improv.$ means the improvement ratio of UPGAN over the strongest baseline. We use ``$A-E$'' to denote the five groups with a decreasing  sparsity level.}
	\label{tab:answer-entity-freq}%
	\begin{small}
		\begin{tabular}{|l|l|c c c c c |}
			\hline
			Dataset&Models&A&B&C&D&E\\
			\hline 
			\hline
			\multirow{4}{*}{Movie}&	DistMult&	10.8&	14.4&	14.6&	22.1&	77.0\\
			&	ConvTransE&	10.4&	13.7&	14.4&	\textbf{22.4}&	76.7\\
			&	UPGAN&	\textbf{13.3}&	\textbf{15.7}&	\textbf{14.8}&	21.6&	\textbf{78.9}\\
			 &	\%Improv. &	+23.1\%&	+9.0\%&	+1.4\%&	-3.6\%&	+2.5\%\\
			\hline
			\multirow{4}{*}{Music}&	DistMult&	73.5&	71.9&	74.2&	72.3&	70.5\\
			&	ConvTransE&	73.8&	72.5&	75.1&	74.1&	73.6\\
			&	UPGAN&	\textbf{76.7}&	\textbf{74.3}&	\textbf{77.0}&	\textbf{75.1}&	\textbf{76.3}\\
			 &	\%Improv. &	+3.9\%&	+2.5\%&	+2.5\%&	+1.3\%&	+3.7\%\\
			\hline
			\multirow{4}{*}{Book}&	DistMult&	8.1&	15.6&	33.7&	47.5&	84.1\\
			&	ConvTransE&	7.3&	15.0&	33.9&	45.8&	82.0\\
			&	UPGAN&	\textbf{8.9}&	\textbf{17.7}&	\textbf{36.4}&	\textbf{52.4}&	\textbf{87.1}\\
			 &	\%Improv. &	+9.9\%&	+13.5\%&	+7.4\%&	+10.3\%&	+3.6\%\\
			\hline
		\end{tabular}%
	\end{small}
\end{table}%
\subsubsection{Performance Comparison w.r.t.  Sparsity Levels }
In KG, different entities correspond to a varying number of triples. Various methods need sufficient training triples for learning good entity representations. Here, we examine how our  method improves over the baseline methods, especially in the sparse case. For this purpose, we first 
divide the test queries into five groups \emph{w.r.t.} the frequency of the answer entity. A smaller group ID indicates that the answer entity of that case occur fewer in training set.
We present the comparison results  in Table~\ref{tab:answer-entity-freq}.
We can see that overall our approach is substantially better than  over baseline methods  in five  sparsity levels. 
Especially, on movie and book datasets, it yields a larger improvement in sparse groups.



\subsubsection{Performance Comparison w.r.t. Hop Number} 
In our dataset, only the aligned KG entities correspond to interaction data from external application systems.   
We have constructed an interaction-augmented KG, and try to learn high-order relatedness between users and entities. 
The preference learned from such high-order relatedness has been verified to be effective in improving the KGC task. 
Hence, we would like to check how the distance of a KG entity to user nodes affects the performance. 
We consider three groups, namely aligned entities (1-hop), attributional entities corresponding to aligned entities (2-hop) and other entities (3-hop and more).
We present the performance comparison of the three groups in Table~\ref{tab:hop-res}.
It can be seen that our method has yielded a substantial improvement in all three groups. 
The finding indicates that our two-stage learning algorithm is able to perform effective information propagation and learning over the heterogeneous graph.
Interestingly,   the 1-hop entities do not always receive the most improvement. Indeed, we have found that the improvement is mainly related to the query difficulty instead of the hop number.  


\begin{table}[htbp]
	\centering
	\caption{Performance (H@3 in precent) comparison \emph{w.r.t.} different hop numbers. $\%Improv.$ means the improvement ratio of UPGAN over the strongest baseline.}
	\label{tab:hop-res}%
	\begin{small}
		\begin{tabular}{|c|c|c c | c c|}
			\hline
			\multirow{2}{*}{Datasets}&\multirow{2}{*}{Hops}&\multicolumn{2}{c|}{Baselines}&\multirow{2}{*}{UPGAN}&\multirow{2}{*}{\%Improv.}\\
			\cline{3-4}
			&&DistMult&ConvTransE&&\\
			\hline \hline
			&	1&	42.8&	43.1&	\textbf{45.0}&	(+4.4\%)\\
			Movie&	2&	8.3& 8.0&	\textbf{8.5}&	(+2.4\%)\\
			&>=3 &	60.7&	59.2&	\textbf{62.0}&	(+2.1\%)\\
			\hline
			&1 &	89.1&	89.6&	\textbf{91.3}&	(+1.9\%)\\
			Music&2 &	71.8&	74.0&	\textbf{76.3}&	(+3.1\%)\\
			&>=3 &	32.5&	33.1&	\textbf{34.9}&	(+5.4\%)\\
			\hline
			&1 &	66.5&	64.2&	\textbf{70.1}&	(+5.4\%)\\
			Book&2 &	16.3&	16.6&	\textbf{18.2}&	(+9.6\%)\\
			&>=3 &	55.5&	53.0&	\textbf{58.7}&	(+5.8\%)\\
			\hline
		\end{tabular}%
	\end{small}
\end{table}%

\subsubsection{Ablation Study}
To effectively utilize the user interaction data, our approach has made several technical extensions. Here, we examine how each of them affects the final performance. 
We consider the following variants of our approach for comparison:

	$\bullet$ \emph{UPGAN$_{\neg G}$}: the variant with only the discriminator component.
	
	$\bullet$ \emph{UPGAN$_{\neg UI}$}: the variant drops the enhanced entity representation from $\bm{x}_q$ (Eq.~\ref{eq-p_h_xq}). In other words, the two-stage learning component has been removed. 
	
	$\bullet$ \emph{UPGAN$_{R-GCN}$}: the variant replaces the two-stage learning component with a neural network architecture similar to R-GCN~\cite{Schlichtkrull-2018-ESWC}. In this variant, we treat all the types of nodes equally.  
	
In  Table~\ref{tab:ablation}, we can see that the performance order can be summarized as:
UPGAN$_{\neg UI}$ $<$ UPGAN$_{R-GCN}$ $<$ UPGAN$_{\neg G}$ $<$  UPGAN.
These results indicate that the proposed techniques are useful to improve the performance. Especially, 
user interaction data with a suitable modeling way is more important for our approach. 

\ignore{First, the results show that removing the generator substantially affects the performance of UPGAN on metric MRR and H@1, which indicates that the negative samples provided by generator can help improve ranking performance.
Second, UPGAN$_{\neg UI}$ and UPGAN$_{R-GCN}$ are worse than the complete UPGAN. The findings indicate that user interaction data is useful to consider in the KGC task. However, it is important to adopt a suitable way to learn and represent such information. 
In a summary, our proposed technical contributions are useful to improve the KGC task by leveraging user interaction data. 
}
\begin{table}[htbp]
	\centering
	\caption{Ablation analysis on the book dataset (in percent).}
	\label{tab:ablation}%
	\begin{small}
		\begin{tabular}{|c |c c c c c|}
			\hline
			Models&MR&MRR&H@1&H@3&H@10\\
			\hline \hline
			UPGAN 	&	\textbf{3463}& \textbf{37.0}& \textbf{30.6}& \textbf{40.5}& \textbf{48.8}\\
			\hline
			UPGAN$_{\neg G}$ & 3546&	36.1&	29.4&	39.8&	48.1\\
			UPGAN$_{R-GCN}$ & 3883& 35.8& 29.8& 39.0& 47.0\\
			UPGAN$_{\neg UI}$ &	5501&	35.0&	28.8&	38.3&	46.7\\
			\hline
		\end{tabular}%
	\end{small}
\end{table}%

\ignore{
\begin{figure}[ht]
	\centering
	\subfigure[Varying the amount of KG triples.]{\label{fig-varing-kg}
		\centering
		\includegraphics[width=0.21\textwidth]{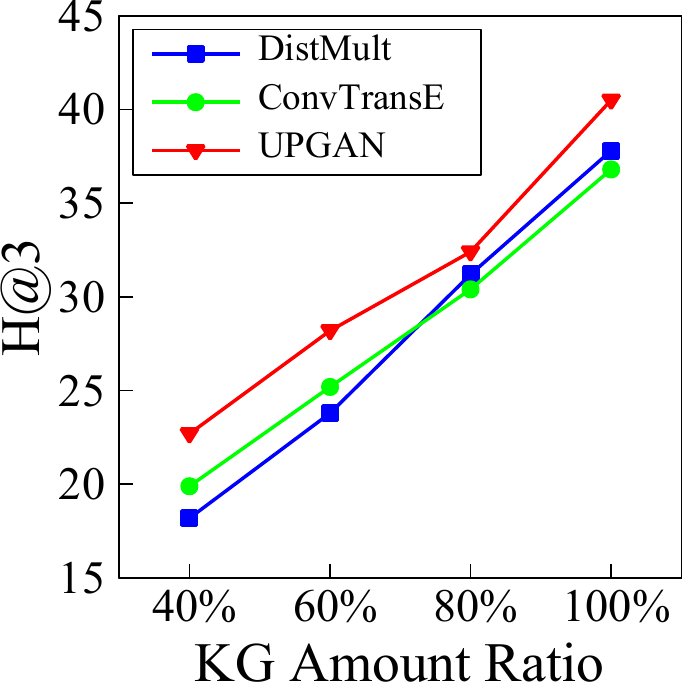}
	}
	\subfigure[Varying the amount of user interaction data.]{\label{fig-varing-ui}
		\centering
		\includegraphics[width=0.21\textwidth]{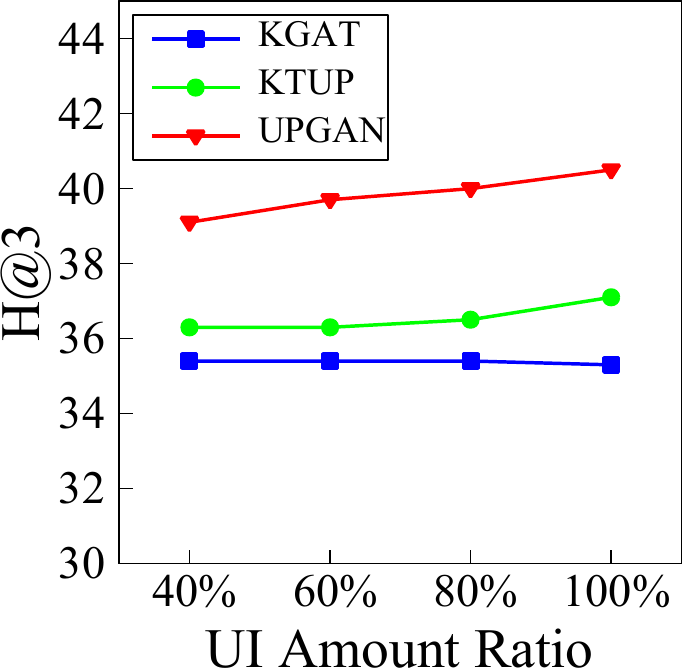}
	}
	\centering
	\caption{Performance tuning  on \textsc{Amazon} book dataset.}
	\label{fig-parameter-tuning}
\end{figure}	
}

\begin{figure}[ht]
 \centering
 \subfigure[Varying the amount of KG triples.]{\label{fig-varing-kg}
  \centering
  \includegraphics[width=0.21\textwidth]{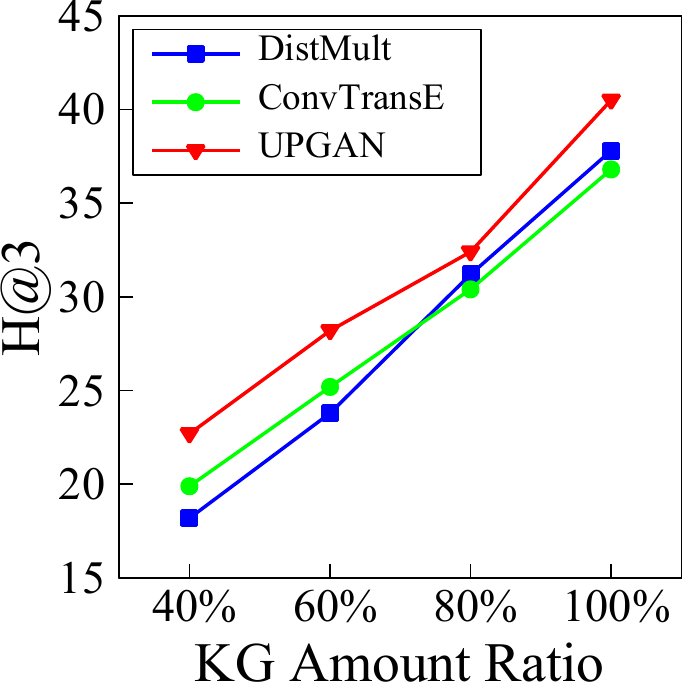}
 }
 \subfigure[Varying the amount of user interaction data.]{\label{fig-varing-ui}
  \centering
  \includegraphics[width=0.21\textwidth]{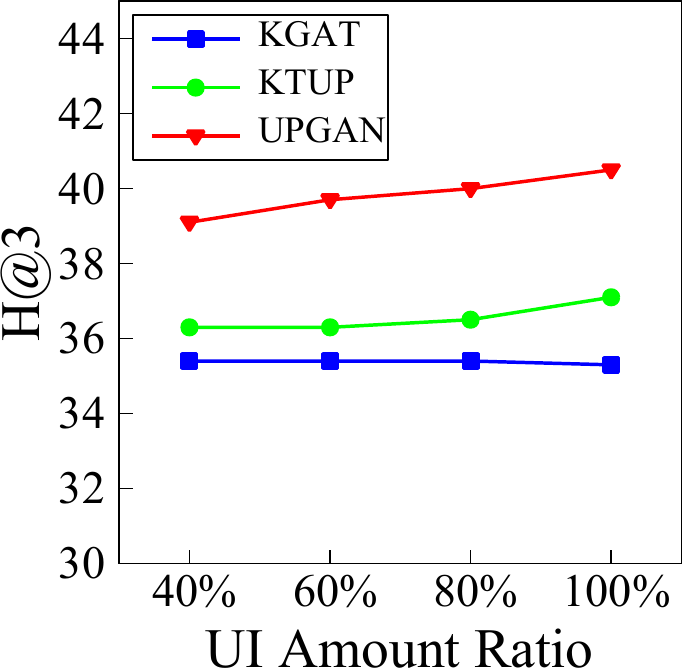}
 }
 \centering
 \caption{Performance tuning on \textsc{Amazon} book dataset.}
 \label{fig-parameter-tuning}
\end{figure}

\subsection{Performance Sensitivity Analysis}

In this part, we further investigate the influence of training data and model parameters on the performance. 
Due to space limit, we only report the results on the book dataset, and omit similar results of the two datasets. 


\begin{figure*}[ht]
	\centering
	\subfigure[Querying the \emph{author} of ``\emph{Part of Bargain}''.]{\label{fig-case-simple}
		\centering
		\includegraphics[height=0.25\textwidth]{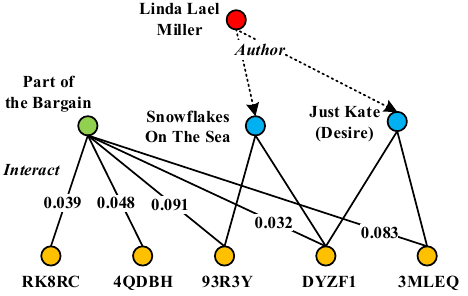}
	}
	\subfigure[Querying the \emph{literary series} of ``\emph{The Riftwar Cycle}''.]{\label{fig-case-complex}
		\centering
		\includegraphics[height=0.3\textwidth]{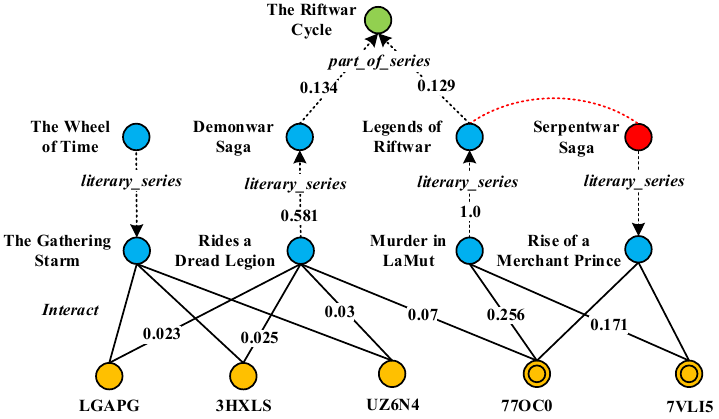}
	}
	\caption{Two cases from \textsc{Amazon} book dataset. We use green, red, blue and yellow circles to denote the target entity, correct entity, KG entity and user respectively. The weights on the edges are computed by our approach. Since the number of the reachable users from the target node is large, we only present five selected users for illustration. }
	\label{fig-case-study}
\end{figure*}

\subsubsection{Varying the amount of KG triples.}
The amount of available KG information directly influences the performance of various KGC methods. Here we examine how our approach performs with the varying amount of KG triples.  
We select DistMult and ConvTransE as comparison methods. We take 40\%, 60\%, 80\% and 100\% from the complete training data to generate four new training sets, respectively. The test set is fixed as original. Fig.~\ref{fig-varing-kg} presents the H@3 performance \emph{w.r.t.} different ratios of KG triples. It can be seen that UPGAN is consistently better than DistMult and ConvTransE with four training sets, especially performs best with an extremely sparse (40\%) amount of KG triples. This observation implies that UPGAN is able to alleviate the influence of data sparsity for KGC methods to some extent. Besides, it can yield more improvement with fewer KG triples. 


\subsubsection{Varying the amount of user interaction data.} 
Since our approach utilizes user interaction data for the KGC task, we continue to examine how its amount affects the final performance. As comparisons, we select two collaborative recommendation and KGC models, namely KGAT and KTUP.
Similarly, we take 40\%, 60\%, 80\% and 100\% from the complete user interaction data to generate four new datasets respectively.
The training set of KG triples and the test set are fixed as original. As we can see from Fig.~\ref{fig-varing-ui}, UPGAN 
is substantially better than KGAT and KTUP for all the four ratios, which indicates the effectiveness of our approach in leveraging user interaction data. Another observation is that the performance of UPGAN gradually increases and the change is relatively stable. 


Besides data amount, we also examine the effect of two parameters, namely the embedding dimensions $K$ and the number of hidden layers in the generator. Overall, we find that it yields a good performance when $K=128$, where the other values in the set  \{16, 32, 64, 128, 256\} give worse results. While, for another parameter, our experiment results show that using two hidden layers give the best performance while the change with other numbers in \{1, 2, 3, 4\} is very small. Due to space limit, we omit the results here. 

\ignore{
\subsubsection{Varying the embedding dimensions}
An important data resource in our model is the trained user and entity embeddings. We now study how different embedding dimensions affect the KGC performance of UPGAN. We vary the user and entity embedding dimension in the set \{16, 32, 64, 128\}. As Fig.~\ref{fig-varing-emb} shows, we can see that an embedding dimension of $64$ gives the best performance for UPGAN. By comparing UPGAN against other baselines in different embedding dimensions, we also observe that UPGAN yields consistently better MRR than the other baselines, which demonstrates the robust performance of UPGAN over varying embedding dimensions. 
\subsubsection{Varying the number of MLP hidden layers in $G$.} In our model, the number of MLP hidden layers in generator $G$ (see Eq.~\ref{generator-mlp}) is an important parameter for learning answer latent representation. Next, we analyze the effects of the number of MLP hidden layers on the MRR performance of UPGAN. As Fig.~\ref{fig-varing-mlp} shows, UPGAN with hidden layer $2$ achieves the best MRR performance. With increasing the number of MLP hidden layers, the performance of UPGAN seems to be similar.
}

\subsection{Case Study}
In this part, we present two cases  for illustrating how our approach utilizes user interaction data for the KGC task.

The first case is related to a query about the \emph{author} for the book ``\emph{Part of Bargain}''.
In our training set, there are few related triples for the book entity ``\emph{Part of Bargain}''.
By only considering KG information, it is difficult for a KGC method to identify the correct answer, since the learned entity representations are not reliable with very limited training data.  
When incorporating the user-item interaction data, we can clearly see that it has several overlapping users with the other two books   ``\emph{Snowflakes On The Sea}'' and ``\emph{Just Kate (Desire)}''. Interestingly, the three related books are written by the same author ``\emph{Linda Lael Miller}''. By running our approach, we can identify the correct answer to this query.

The second case is related to a query about the relation \emph{part~\_of~\_series} for the book series, which aims to identify the literary series (\aka sub-series) that belong to ``\emph{The Riftwar Cycle}'' (target entity).
Following the first case, we check whether the related users on the graph can be useful for this query.
Starting from the target entity, we can identify 128 related users in total with the BFS extension based on the
interaction-augmented KG. 
Given two candidate literary series ``\emph{Serpentwar Saga}'' and ``\emph{The Wheel of Time}'', a straightforward method is to count the number of a literary series that has been read by the related users. 
However, ``\emph{The Wheel of Time}'' is much more popular than the correct entity ``\emph{Serpentwar Saga}'' (33 \emph{v.s.} 17).
It indicates that simply using the user interaction data may incorporate noise. 
As a comparison, by running our approach, we can identify more important users on the graph. 
As we can see, the two users with ID ``77OC0'' and ``7VLI5'' are assigned with very large attention weights by our algorithm. An interesting observation is that ``\emph{Legends of the Riftwar}'' and ``\emph{Serpentwar Saga}'' can be associated via the two selected users. 
Based on the known fact that ``\emph{Legends of the Riftwar}'' belongs to ``\emph{The Riftwar Cycle}'', our approach is capable of identifying ``\emph{Serpentwar Saga}'' as the final answer.


%% file: sec-con.tex
\section{Conclusion}

In this paper, we developed an adversarial learning approach for effectively learning useful information from user interaction data for the KGC task. Especially, we have made three major technical contributions. First,  we  constructed an interaction-augmented KG for unifying KG and user interaction data, and design a two-stage representation learning algorithm for collaboratively learning effective representations for heterogeneous nodes. Second, by integrating enhanced entity representations, we designed a user preference guided discriminator for evaluating the plausibility of a candidate entity given a query. 
Third, we designed a query-specific generator for producing hard negative entities for given a query.
We constructed evaluation experiments with three large datasets. The results showed that our proposed model is superior to previous methods in terms of effectiveness for the KGC task. 

Currently, only three datasets with aligned entity-item linkage  have been used for evaluation. We believe our approach is applicable to more domains. 
In the future, we will investigate into how our models perform in other domains. 

